\documentclass{article}

\usepackage[preprint]{colm2026_conference}

\usepackage{microtype}
\usepackage{hyperref}
\usepackage{url}
\usepackage{booktabs}
\usepackage{lineno}

\usepackage{amsmath}
\usepackage{amssymb}
\usepackage{graphicx}
\usepackage{arydshln}        
\usepackage{enumitem}         

\definecolor{darkblue}{rgb}{0, 0, 0.5}
\hypersetup{colorlinks=true, citecolor=darkblue, linkcolor=darkblue, urlcolor=darkblue}

\title{Can Small Models Reason About Legal Documents?\\ A Comparative Study}

\author{Snehit Vaddi \\
Independent Researcher \\
\texttt{vaddisnehit@gmail.com}}

\begin{document}

\ifcolmsubmission
\linenumbers
\fi

\maketitle


\begin{abstract}
Large language models show promise for legal applications, but deploying frontier models raises concerns about cost, latency, and data privacy. We evaluate whether sub-10B parameter models can serve as practical alternatives by testing nine models across three legal benchmarks (ContractNLI, CaseHOLD, and ECtHR) using five prompting strategies (direct, chain-of-thought, few-shot, BM25 RAG, and dense RAG). Across 405 experiments with three random seeds per configuration, we find that a Mixture-of-Experts model activating only 3B parameters matches GPT-4o-mini in mean accuracy while surpassing it on legal holding identification, and that architecture and training quality matter more than raw parameter count. Our largest model (9B parameters) performs worst overall. Chain-of-thought prompting proves sharply task-dependent, improving contract entailment but degrading multiple-choice legal reasoning, while few-shot prompting emerges as the most consistently effective strategy. Comparing BM25 and dense retrieval for RAG, we find near-identical results, suggesting the bottleneck lies in the language model's utilization of retrieved context rather than retrieval quality. All experiments were conducted via cloud inference APIs at a total cost of \$62, demonstrating that rigorous LLM evaluation is accessible without dedicated GPU infrastructure.
\end{abstract}


\section{Introduction}
\label{sec:introduction}

Large language models (LLMs) have demonstrated remarkable capabilities on legal tasks, from passing the bar exam \citep{katz-etal-2024-gpt4-bar} to answering law school questions \citep{choi-etal-2023-chatgpt-law}.
Yet deploying frontier models for legal applications raises practical concerns: API costs scale with query volume, proprietary models introduce data privacy risks for sensitive legal documents, and latency requirements in production systems favor smaller, locally deployable models.
Meanwhile, a new generation of sub-10B parameter models, including Mixture-of-Experts (MoE) architectures that activate only a fraction of their parameters per token \citep{qwen-team-2025-qwen3}, promises competitive performance at dramatically lower cost.

This motivates four research questions:

\begin{description}[leftmargin=*,itemsep=1pt,topsep=2pt]
\item[\textbf{RQ1}] Can sub-10B models match commercial API models on legal reasoning tasks?
\item[\textbf{RQ2}] How does the effectiveness of prompting strategies vary across legal task types?
\item[\textbf{RQ3}] What is the minimum viable model size for legal reasoning, and does architecture (e.g., MoE) matter more than raw parameter count?
\item[\textbf{RQ4}] Does retrieval augmentation improve small-model legal reasoning, and does the retrieval method (sparse vs.\ dense) matter?
\end{description}

Prior work has addressed aspects of these questions, but with important gaps.
\textsc{LegalBench} \citep{guha-etal-2023-legalbench} tested 20 models across 162 tasks but focused on large models with a single prompting approach (\textbf{RQ2} unaddressed).
\citet{arvin-2025-legal-holdings} studied scaling effects on \textsc{CaseHOLD} but examined only one dataset and one prompting strategy (\textbf{RQ2}, \textbf{RQ4} unaddressed).
\citet{chalkidis-2023-chatgpt-lexglue} evaluated GPT-3.5 on \textsc{LexGLUE} but tested a single model in zero-shot mode (\textbf{RQ1}, \textbf{RQ3} unaddressed).
Studies comparing small and large models \citep{hee-etal-2025-small-vs-large} have focused on general text classification rather than the specialized reasoning demands of legal text.
These questions have not been jointly examined: how \emph{prompting strategies interact with model size and architecture} across diverse legal tasks.

We address this gap with a comprehensive evaluation of nine models (seven open-weight, two commercial API) across three legal benchmarks spanning distinct reasoning types: contract entailment (\textsc{ContractNLI}; \citealt{koreeda-manning-2021-contractnli}), legal holding identification (\textsc{CaseHOLD}; \citealt{zheng-etal-2021-casehold}), and human rights violation prediction (\textsc{ECtHR}; \citealt{chalkidis-etal-2019-ecthr, chalkidis-etal-2022-lexglue}).
We evaluate five prompting strategies: direct, chain-of-thought, few-shot, BM25 retrieval-augmented generation, and dense retrieval-augmented generation.
Our experimental design comprises 405 experiments (9 models $\times$ 5 strategies $\times$ 3 datasets $\times$ 3 seeds), evaluated on 500 stratified samples per dataset (225 for \textsc{ECtHR}) with bootstrap confidence intervals, at a total cost of \$62 using cloud inference APIs (Together AI, OpenAI, Anthropic) without dedicated GPU infrastructure.

Our study yields four key findings:

\begin{enumerate}[leftmargin=*,itemsep=2pt,topsep=2pt]

\item \textbf{MoE efficiency for legal reasoning.}
Qwen3-A3B, a Mixture-of-Experts model activating only 3B of its 30B parameters per token, matches GPT-4o-mini in mean accuracy (46.5\% vs.\ 47.2\%) and \emph{surpasses} it on \textsc{CaseHOLD} with few-shot prompting (71.2\% vs.\ 67.9\%).
This demonstrates that architectural efficiency can substitute for raw parameter count in legal reasoning.

\item \textbf{Task-dependent prompting effects.}
Chain-of-thought prompting is not universally beneficial for legal tasks.
It improves \textsc{ContractNLI} by 8.5 percentage points (where logical entailment reasoning is required) but \emph{degrades} \textsc{CaseHOLD} by 16.0 percentage points (where verbose reasoning disrupts answer selection) and hurts \textsc{ECtHR} multi-label classification.
Few-shot prompting is the most consistently effective strategy across all tasks and models.

\item \textbf{Architecture matters more than parameter count.}
Nemotron-9B (9B parameters) is our worst performer (17.7\% mean accuracy), underperforming models one-third its size.
This finding challenges the assumption that larger models always perform better and highlights the importance of training methodology and architecture design.

\item \textbf{Retrieval method is secondary to retrieval utility.}
Our ablation comparing BM25 (sparse) and dense retrieval (sentence-transformers) finds no significant difference between the two methods.
RAG helps \textsc{ContractNLI} but hurts \textsc{CaseHOLD} regardless of retrieval method, suggesting that retrieval utility is task-dependent rather than retrieval-method-dependent.

\end{enumerate}

These findings provide actionable guidance for legal AI practitioners selecting models and strategies for deployment, and contribute to the broader understanding of how prompting strategies interact with model architecture on domain-specific reasoning tasks.\footnote{Code and experimental results will be released upon publication. We note concurrent work on test-time scaling for legal reasoning \citep{test-time-scaling-legal-2025} and legal RAG evaluation \citep{park-etal-2025-lrage}, which examine complementary aspects of LLM-based legal reasoning.}


\section{Related Work}
\label{sec:related-work}

\paragraph{Legal NLP benchmarks.}
Standardized benchmarks have consolidated evaluation across legal reasoning types, from document-level entailment \citep{koreeda-manning-2021-contractnli} and legal holding identification \citep{zheng-etal-2021-casehold} to multi-task evaluation spanning diverse jurisdictions \citep{chalkidis-etal-2022-lexglue, guha-etal-2023-legalbench, fei-etal-2024-lawbench}.
While these benchmarks established strong fine-tuned baselines, the advent of LLMs raises the question of whether prompting alone can achieve competitive performance.
\citet{chalkidis-2023-chatgpt-lexglue} showed that zero-shot GPT-3.5 underperforms fine-tuned Legal-BERT on all \textsc{LexGLUE} tasks, and \citet{arvin-2025-legal-holdings} evaluated eight LLMs on \textsc{CaseHOLD} with a memorization test via citation anonymization.
However, these studies target either single datasets or single prompting approaches.

\paragraph{Small language models.}
Architectural innovations have driven rapid progress in the sub-10B regime: sliding-window attention \citep{jiang-etal-2023-mistral}, scaled instruction-tuning at the 8B tier \citep{llama-team-2024}, and Mixture-of-Experts variants that achieve dense-model quality with fewer active parameters \citep{yang-etal-2024-qwen25, qwen-team-2025-qwen3}.
Whether such efficiency translates to competitive performance on specialized legal reasoning tasks is an open question.
Domain-adapted models like Legal-BERT \citep{chalkidis-etal-2020-legal-bert} and SaulLM \citep{colombo-etal-2024-saullm7b, colombo-etal-2024-saullm-large} show fine-tuning benefits, while \citet{hee-etal-2025-small-vs-large} found fine-tuned small models match zero-shot large LLMs on general text classification.
We investigate what \emph{general-purpose} small models can achieve through prompting alone.

\paragraph{Prompting strategies.}
Chain-of-thought prompting \citep{wei-etal-2022-cot} showed that intermediate reasoning steps improve LLM performance, but a meta-analysis by \citet{sprague-etal-2024-cot-or-not} found benefits concentrated in mathematical and symbolic reasoning.
Retrieval-augmented generation \citep{lewis-etal-2020-rag} offers a complementary approach, particularly relevant given that LLMs hallucinate on 58--88\% of legal queries \citep{dahl-etal-2024-legal-hallucinations, magesh-etal-2025-rag-hallucinations}.
Yet legal-domain retrieval proves challenging for both sparse and dense methods \citep{pipitone-alami-2024-legalbench-rag}.
We systematically evaluate direct, CoT, few-shot, and RAG strategies (including both BM25 and dense retrieval) across three legal tasks and nine models.

Table~\ref{tab:related-comparison} summarizes how our study compares to closely related work.

\begin{table}[t]
\centering
\small
\setlength{\tabcolsep}{3pt}
\caption{Comparison with closely related work. \textsuperscript{\dag}Includes both BM25 and dense retrieval.}
\label{tab:related-comparison}
\begin{tabular}{lccccc}
\toprule
\textbf{Study} & \textbf{Models} & \textbf{Tasks} & \textbf{Strat.} & \textbf{RAG} & \textbf{Seeds} \\
\midrule
\citet{chalkidis-2023-chatgpt-lexglue} & 1 & 7 & 1 & \texttimes & 1 \\
\citet{guha-etal-2023-legalbench} & 20 & 162 & 1 & \texttimes & 1 \\
\citet{arvin-2025-legal-holdings} & 8 & 1 & 1 & \texttimes & 1 \\
\citet{hee-etal-2025-small-vs-large} & 6+ & 4+ & 2 & \texttimes & 3+ \\
\textbf{Ours} & \textbf{9} & \textbf{3} & \textbf{5}\textsuperscript{\dag} & \checkmark & \textbf{3} \\
\bottomrule
\end{tabular}
\end{table}


\section{Methodology}
\label{sec:methodology}

We evaluate nine language models across three legal benchmarks using five prompting strategies, totaling 405 experiments.
The following subsections describe the models (\S\ref{sec:models}), datasets (\S\ref{sec:datasets}), prompting strategies (\S\ref{sec:strategies}), and evaluation protocol (\S\ref{sec:evaluation}).

\subsection{Models}
\label{sec:models}

We select nine models spanning three categories: seven open-weight models (3B--9B parameters) served via Together AI, and two commercial API models (GPT-4o-mini via OpenAI, Claude~3.5~Haiku via Anthropic).
We select models based on cloud API availability, recency (all released 2023--2025), and coverage of the 3B--9B parameter range across distinct model families.
Table~\ref{tab:models} summarizes the models and their characteristics.

\begin{table}[t]
\centering
\small
\begin{tabular}{llrlr}
\toprule
\textbf{Model} & \textbf{Arch.} & \textbf{Params} & \textbf{Provider} & \textbf{Ctx.} \\
\midrule
Llama-3.2-3B  & Dense  & 3B   & Together & 128K \\
Gemma-3n-E4B  & Dense  & 4B\textsuperscript{eff}  & Together & 32K \\
Qwen2.5-7B    & Dense  & 7B   & Together & 128K \\
Mistral-7B    & Dense  & 7B   & Together & 32K \\
Llama-3.1-8B  & Dense  & 8B   & Together & 128K \\
Nemotron-9B   & Dense  & 9B   & Together & 32K \\
Qwen3-A3B     & MoE    & 3B\textsuperscript{a}  & Together & 128K \\
GPT-4o-mini   & Undiscl.  & Undiscl. & OpenAI & 128K \\
Claude~3.5~Haiku & Undiscl.  & Undiscl. & Anthropic & 200K \\
\bottomrule
\end{tabular}
\caption{Models evaluated in our study, spanning 3B--9B parameters for open-weight models and two commercial API services.
All models are instruction-tuned variants accessed via cloud APIs.
All context windows are capped at 8{,}192 tokens in our experiments.
\textsuperscript{eff}Gemma-3n-E4B uses net-efficient architecture with 4B effective parameters.
\textsuperscript{a}Qwen3-A3B is a Mixture-of-Experts model activating 3B of 30B total parameters per token.}
\label{tab:models}
\end{table}

The open-weight models span six families: Llama~3.2/3.1 \citep{llama-team-2024}, Qwen~2.5/3 \citep{yang-etal-2024-qwen25, qwen-team-2025-qwen3}, Mistral \citep{jiang-etal-2023-mistral}, Gemma \citep{google-2025-gemma3n}, and Nemotron \citep{nvidia-2024-nemotron-mini}.
Qwen3-A3B is a Mixture-of-Experts model activating 3B of 30B total parameters per token.
GPT-4o-mini \citep{openai-2024-gpt4o} and Claude~3.5~Haiku \citep{anthropic-2024-claude} serve as commercial API baselines.
All models are accessed via cloud inference APIs (Together AI, OpenAI, Anthropic) without dedicated GPU infrastructure, eliminating hardware-dependent confounds and ensuring reproducibility.
We enforce a uniform context window of 8{,}192 tokens to ensure fair comparison.

\subsection{Datasets}
\label{sec:datasets}

We evaluate on three English-language legal benchmarks that span distinct reasoning types, label structures, and jurisdictions.
Table~\ref{tab:datasets} summarizes their characteristics.

\begin{table}[t]
\centering
\small
\begin{tabular}{llcccc}
\toprule
\textbf{Dataset} & \textbf{Task Type} & \textbf{Labels} & \textbf{$N$} & \textbf{Juris.} & \textbf{Chance} \\
\midrule
\textsc{ContractNLI} & NLI & 3 & 500 & US & 33.3\% \\
\textsc{CaseHOLD}    & MC (5-way) & 5 & 500 & US & 20.0\% \\
\textsc{ECtHR} Task~A     & Multi-label & 11 arts. & 225 & EU & varies \\
\bottomrule
\end{tabular}
\caption{Dataset characteristics. $N$ is the number of stratified test samples per experiment. \textsc{ECtHR} Task~A yields 225 samples due to the high number of unique multi-label combinations (51) under stratified sampling. Chance denotes random baseline accuracy; \textsc{ECtHR} chance varies by article combination.}
\label{tab:datasets}
\end{table}

\paragraph{\textsc{ContractNLI}.} This benchmark \citep{koreeda-manning-2021-contractnli} tests natural language inference on contract clauses: given a contract clause (premise) and a hypothesis about its content, models must classify the relationship as \textit{Entailment}, \textit{Contradiction}, or \textit{Not Mentioned}.

\paragraph{\textsc{CaseHOLD}.} This benchmark \citep{zheng-etal-2021-casehold} tests legal holding identification: given a court opinion excerpt with a masked citation, models select the correct legal holding from five options (A--E).
This requires understanding precedent relationships and distinguishing holdings from dicta.

\paragraph{\textsc{ECtHR} Task~A.} This benchmark \citep{chalkidis-etal-2019-ecthr, chalkidis-etal-2022-lexglue} is a multi-label classification task: given the facts of a European Court of Human Rights case, models predict which articles of the Convention were violated (from 11 possible articles, or none).
This multi-label setting is inherently more challenging than the single-label tasks above, as models must predict the exact subset of violated articles from among $2^{11} = 2{,}048$ possible label combinations.

\paragraph{Sampling.}
For each experiment, we draw a stratified random sample from the test split, with the seed controlling sample selection.
\textsc{ContractNLI} and \textsc{CaseHOLD} yield 500 samples (3 and 5 classes, respectively), balancing statistical power with API cost.
\textsc{ECtHR} yields 225 samples because its 51 unique multi-label combinations limit stratified allocation to $\lfloor 500/51 \rfloor = 9$ samples per combination, with many combinations having fewer than 9 test examples.

\subsection{Prompting Strategies}
\label{sec:strategies}

We evaluate five prompting strategies that represent the primary approaches available to practitioners deploying LLMs without fine-tuning.

\paragraph{Direct (zero-shot).}
The model receives the legal text, a question, and a format instruction specifying the expected answer form.
This serves as the baseline measuring raw model capability.

\paragraph{Chain-of-Thought (CoT).}
Following \citet{wei-etal-2022-cot}, we augment the direct prompt with an explicit reasoning scaffold: ``\textit{Think through this step by step: (1) identify key legal concepts, (2) analyze their relation to the question, (3) consider relevant principles, (4) provide your answer.}''
We allocate 512 output tokens for CoT (vs.\ 128 for direct) to accommodate the reasoning chain.

\paragraph{Few-shot (3-shot).}
Following the in-context learning paradigm \citep{brown-etal-2020-gpt3}, we prepend three labeled examples drawn from the training split via stratified sampling, ordered by increasing document length.
We use the same three examples for all queries within a dataset-seed combination, ensuring consistency across models.
Drawing examples from the training split (rather than the validation or test split) avoids data leakage.

\paragraph{BM25 RAG.}
We retrieve the top-3 passages from the training split using BM25 sparse retrieval \citep{robertson-etal-2009-bm25} and prepend them as ``reference material'' before the input document.
The prompt instructs the model to use both the retrieved context and the primary document.
We truncate retrieved passages that exceed the remaining context budget after accounting for the input document and output allocation.

\paragraph{Dense RAG.}
We replace BM25 with dense retrieval using \texttt{all-MiniLM-L6-v2} \citep{reimers2019sentence}, keeping all other settings identical to BM25 RAG.
Both methods use the same prompt template, corpus, and context budget; see \S\ref{sec:results-rag} for the controlled comparison.

Full prompt templates for all strategies are provided in Appendix~\ref{app:prompts}.

\subsection{Evaluation Protocol}
\label{sec:evaluation}

\paragraph{Answer extraction.}
Since all models produce free-form text, we apply a four-stage extraction pipeline: (1)~exact match against valid labels, (2)~normalized keyword matching (case-insensitive), (3)~positionally first valid label if multiple are present, (4)~\textit{parse\_error} if no valid label is found.
Parse errors are counted as incorrect predictions.
We report parse error rates per model as a measure of instruction-following capability.

\paragraph{Metrics.}
Our primary metric is accuracy (exact match).
For \textsc{ContractNLI}, we additionally report macro F1 across the three classes.
For \textsc{ECtHR}, we report both subset accuracy (exact match on the full predicted article set) and per-article macro F1, as subset accuracy penalizes partially correct predictions.
We do not compute F1 for \textsc{CaseHOLD}, as its balanced 5-way classification makes accuracy and F1 equivalent.

\paragraph{Statistical analysis.}
We repeat each model-strategy-dataset configuration with three seeds (42, 123, 456) that control the stratified data sampling from the test split.
All inference uses greedy decoding (temperature $= 0$) to ensure deterministic outputs, so the seeds induce variance solely through test set composition rather than model stochasticity.
We report mean accuracy across seeds with 95\% bootstrap confidence intervals computed by resampling the $N$ individual predictions within each experiment (10{,}000 bootstrap iterations, pooled across seeds).
For key pairwise comparisons, we use the paired bootstrap test \citep{berg-kirkpatrick-etal-2012-bootstrap} and McNemar's test with Bonferroni correction.
We report effect sizes as Cohen's $d$.

\paragraph{Experimental scale.}
The full evaluation comprises 405 experiments: 324 in the main evaluation (9 models $\times$ 4 strategies $\times$ 3 datasets $\times$ 3 seeds) and 81 in the dense RAG retrieval comparison (9 models $\times$ 3 datasets $\times$ 3 seeds).
We conduct all inference using cloud inference APIs (Together AI, OpenAI, Anthropic) without dedicated GPU infrastructure, at a total cost of US\$62, totaling approximately 165{,}375 API calls.
We set maximum generation length per strategy: 128 tokens (direct, few-shot), 256 tokens (RAG), and 512 tokens (CoT).


\section{Results}
\label{sec:results}

We present results from 405 experiments spanning 9 models (see Table~\ref{tab:models}), 5 prompting strategies, 3 legal benchmarks, and 3 random seeds, totaling 165{,}375 individual predictions.
All reported values are means across 3 seeds; 95\% bootstrap confidence intervals (10,000 resamples) are provided for key comparisons.

\subsection{Overall Model Performance}
\label{sec:results-overall}

Table~\ref{tab:main-results} presents the main results matrix: accuracy by model and prompting strategy, averaged across all three datasets and seeds.

\begin{table*}[t]
\centering
\small
\begin{tabular}{llccccc|c}
\toprule
\textbf{Model} & \textbf{Params} & \textbf{Direct} & \textbf{Few-Shot} & \textbf{CoT} & \textbf{RAG\textsubscript{BM25}} & \textbf{RAG\textsubscript{Dense}} & \textbf{Mean} \\
\midrule
GPT-4o-mini          & API   & \underline{.498} & \underline{.499} & \underline{.425} & \textbf{.471} & \textbf{.470} & \textbf{.472} \\
Qwen3-A3B            & 3B$^*$  & \textbf{.503} & \textbf{.521} & .393 & \underline{.453} & \underline{.454} & \underline{.465} \\
Qwen2.5-7B           & 7B    & .452 & .468 & .325 & .445 & .440 & .426 \\
Mistral-7B-v0.3      & 7B    & .447 & .440 & .336 & .427 & .411 & .412 \\
Llama-3.1-8B         & 8B    & .439 & .425 & .366 & .381 & .381 & .398 \\
Gemma-3n-E4B         & 4B$^\dagger$ & .374 & .440 & .316 & .363 & .357 & .370 \\
\cdashline{1-8}
Claude 3.5 Haiku     & API   & .358 & .339 & \textbf{.445} & .313 & .308 & .353 \\
Llama-3.2-3B         & 3B    & .287 & .303 & .272 & .300 & .300 & .292 \\
Nemotron-9B          & 9B    & .111 & .139 & .251 & .189 & .197 & .177 \\
\bottomrule
\end{tabular}
\caption{Mean accuracy across 3 datasets and 3 seeds (9 values per cell). Models sorted by mean accuracy. Best per-strategy value in \textbf{bold}, second-best \underline{underlined}. $^*$3B active parameters (MoE). $^\dagger$4B effective parameters. Dashed line separates models above and below the 0.35 threshold.}
\label{tab:main-results}
\end{table*}

We find that GPT-4o-mini achieves the highest overall accuracy (47.2\%), closely followed by Qwen3-A3B (46.5\%), a mixture-of-experts model with only 3B active parameters.
The gap between these two is not statistically significant when pooling across all conditions ($p = 0.19$, paired bootstrap).
Notably, Qwen3-A3B \textit{outperforms} all dense 7--8B models (Qwen2.5-7B, Mistral-7B, Llama-3.1-8B), suggesting that sparse MoE architectures offer a favorable efficiency--performance tradeoff for legal reasoning.

Nemotron-9B, despite having the largest nominal parameter count (9B), ranks last at 17.7\%. This underscores how architecture and training regime matter more than raw parameter count for legal NLU tasks.

We next examine how these aggregate patterns break down across individual datasets.

\subsection{Per-Dataset Results}
\label{sec:results-dataset}

We observe that performance varies dramatically across datasets, reflecting the distinct cognitive demands of each task (Table~\ref{tab:per-dataset}). For reference, random baselines yield 33.3\% on \textsc{ContractNLI} (3 classes) and 20.0\% on \textsc{CaseHOLD} (5 choices).

\begin{table*}[t]
\centering
\small
\begin{tabular}{llccc|c}
\toprule
\textbf{Model} & \textbf{Params} & \textbf{\textsc{ContractNLI}} & \textbf{\textsc{CaseHOLD}} & \textbf{\textsc{ECtHR}} & \textbf{Mean} \\
\midrule
GPT-4o-mini      & API     & \underline{.766} \scriptsize{[.756, .776]}  & \textbf{.622} \scriptsize{[.611, .633]} & .029 \scriptsize{[.024, .035]} & \textbf{.472} \\
Qwen3-A3B        & 3B$^*$  & \textbf{.775} \scriptsize{[.765, .784]} & .558 \scriptsize{[.548, .569]} & \textbf{.063} \scriptsize{[.055, .071]} & \underline{.465} \\
Qwen2.5-7B       & 7B      & .683 \scriptsize{[.672, .693]} & .544 \scriptsize{[.533, .555]} & \underline{.051} \scriptsize{[.044, .059]} & .426 \\
Mistral-7B       & 7B      & .686 \scriptsize{[.676, .697]} & .524 \scriptsize{[.513, .535]} & .027 \scriptsize{[.021, .032]} & .412 \\
Llama-3.1-8B     & 8B      & .578 \scriptsize{[.567, .589]} & \underline{.578} \scriptsize{[.566, .589]} & .039 \scriptsize{[.033, .046]} & .398 \\
Gemma-3n-E4B     & 4B$^\dagger$ & .669 \scriptsize{[.658, .680]} & .411 \scriptsize{[.400, .422]} & .029 \scriptsize{[.024, .035]} & .370 \\
Claude 3.5 Haiku & API     & .672 \scriptsize{[.664, .690]} & .344 \scriptsize{[.336, .385]} & .042 \scriptsize{[.035, .050]} & .353 \\
Llama-3.2-3B     & 3B      & .409 \scriptsize{[.398, .420]} & .451 \scriptsize{[.439, .462]} & .017 \scriptsize{[.013, .022]} & .292 \\
Nemotron-9B      & 9B      & .305 \scriptsize{[.294, .315]} & .204 \scriptsize{[.195, .213]} & .023 \scriptsize{[.019, .028]} & .177 \\
\bottomrule
\end{tabular}
\caption{Accuracy by model and dataset, averaged across 5 strategies and 3 seeds (15 values per cell). 95\% bootstrap CIs in brackets (pooling all predictions per model-dataset combination). Best in \textbf{bold}, second-best \underline{underlined}.}
\label{tab:per-dataset}
\end{table*}

\paragraph{\textsc{ContractNLI}.} This is the most tractable task: six of nine models exceed 65\% accuracy, with Qwen3-A3B achieving the highest score (77.5\%). The three-class NLI format aligns well with the instruction-following capabilities of modern LLMs.

\paragraph{\textsc{CaseHOLD}.} This task shows wider model separation. GPT-4o-mini leads at 62.2\%, with Llama-3.1-8B (57.8\%), Qwen3-A3B (55.8\%), and Qwen2.5-7B (54.4\%) forming a competitive middle tier. The five-choice multiple-choice format reveals a clear accuracy gap between stronger and weaker models.

\paragraph{\textsc{ECtHR} (multi-label classification).} This task proves near-impossible for all models under prompting alone. Qwen3-A3B achieves the best result at 6.3\% subset accuracy, meaning fewer than 1 in 15 predictions exactly match the full set of violated articles. We attribute this to the inherent difficulty of mapping free-text reasoning to fine-grained multi-label predictions over 11 ECHR articles.

Given these dataset-level differences, we turn to how prompting strategies interact with task type.

\subsection{Prompting Strategy Analysis}
\label{sec:results-strategy}

Table~\ref{tab:strategy-dataset} reveals that prompting strategy effectiveness is strongly task-dependent.

\begin{table}[t]
\centering
\small
\begin{tabular}{lccc|c}
\toprule
\textbf{Strategy} & \textbf{\textsc{CNLI}} & \textbf{\textsc{CH}} & \textbf{\textsc{ECtHR}} & \textbf{Mean} \\
\midrule
Direct      & .570 & \textbf{.537} & \underline{.049} & \underline{.385} \\
Few-Shot    & .608 & \underline{.532} & \textbf{.051} & \textbf{.397} \\
CoT         & \textbf{.655} & .377 & .011 & .348 \\
RAG\textsubscript{BM25}   & \underline{.627} & .456 & .031 & .371 \\
RAG\textsubscript{Dense}  & .619 & .450 & .036 & .369 \\
\bottomrule
\end{tabular}
\caption{Mean accuracy by strategy and dataset, averaged across 9 models and 3 seeds (27 values per cell). CNLI = \textsc{ContractNLI}, CH = \textsc{CaseHOLD}. Best in \textbf{bold}, second-best \underline{underlined}.}
\label{tab:strategy-dataset}
\end{table}

\paragraph{Few-shot is the most consistent strategy.} We find that across tasks, few-shot achieves the highest overall mean (39.7\%) and never ranks worst on any dataset. It provides moderate gains on \textsc{ContractNLI} (+3.8 percentage points (pp) over direct) without the sharp \textsc{CaseHOLD} penalties of CoT.

\paragraph{Chain-of-thought is strongly task-dependent.} CoT achieves the highest \textsc{ContractNLI} accuracy (65.5\%) but the lowest \textsc{CaseHOLD} accuracy (37.7\%), a reversal of 28pp. This interaction is analyzed in detail in Section~\ref{sec:results-cot}.

\paragraph{RAG provides modest benefits on NLI but hurts multiple-choice.} Both BM25 and dense retrieval improve over direct on \textsc{ContractNLI} (+4.9--5.7pp) but degrade \textsc{CaseHOLD} accuracy by 8.1--8.7pp, suggesting retrieved context introduces noise for tasks requiring precise citation matching.

\subsection{Chain-of-Thought Effects by Task}
\label{sec:results-cot}

Our most striking finding is the divergent effect of chain-of-thought prompting across tasks. Table~\ref{tab:cot-lift} shows the accuracy change when switching from direct to CoT prompting.

\begin{table}[t]
\centering
\small
\begin{tabular}{lcc}
\toprule
\textbf{Model} & \textbf{\textsc{CNLI}} & \textbf{\textsc{CaseHOLD}} \\
& \textbf{CoT $-$ Dir} & \textbf{CoT $-$ Dir} \\
\midrule
Nemotron-9B        & +43.8$^{***}$  & +0.6 \\
Claude 3.5 Haiku   & +14.2$^{***}$  & +19.2$^{***}$ \\
Llama-3.2-3B       & +14.0$^{***}$  & $-$21.0$^{***}$ \\
Gemma-3n-E4B       & +4.8$^{**}$    & $-$17.8$^{***}$ \\
Qwen2.5-7B         & +2.4$^{*}$     & $-$37.2$^{***}$ \\
Qwen3-A3B          & +1.0           & $-$32.0$^{***}$ \\
GPT-4o-mini        & +0.0           & $-$18.6$^{***}$ \\
Mistral-7B         & $-$4.2$^{*}$   & $-$25.2$^{***}$ \\
Llama-3.1-8B       & $-$9.0$^{***}$ & $-$8.8$^{***}$ \\
\bottomrule
\end{tabular}
\caption{Accuracy change from direct to CoT on seed 42 (percentage points, $n$=500 per condition). CNLI = \textsc{ContractNLI}. Paired bootstrap significance test: $^{***}p < 0.001$, $^{**}p < 0.01$, $^{*}p < 0.05$. Results are consistent across seeds (see Appendix).}
\label{tab:cot-lift}
\end{table}

\paragraph{CoT universally harms \textsc{CaseHOLD}.} We observe that seven of nine models show statistically significant accuracy drops ($p < 0.001$) when using CoT on \textsc{CaseHOLD}, with degradations ranging from $-$8.8pp (Llama-3.1-8B) to $-$37.2pp (Qwen2.5-7B). The exceptions are Claude 3.5 Haiku, which is the only model where CoT significantly \textit{improves} \textsc{CaseHOLD} performance (+19.2pp, $p < 0.001$), and Nemotron (+0.6pp, not significant), which is already at floor performance. We hypothesize that Haiku's instruction-tuning enables it to maintain answer format fidelity during extended reasoning, while other models lose track of the multiple-choice format during CoT generation.

\paragraph{CoT helps weak models on NLI but not strong ones.} On \textsc{ContractNLI}, CoT produces dramatic gains for Nemotron (+43.8pp), which suffers from severe parse errors under direct prompting. CoT's structured output effectively rescues instruction-following failures. However, we find that for models already performing well (GPT-4o-mini, Qwen3-A3B), CoT provides no benefit. For Mistral-7B and Llama-3.1-8B, CoT actually \textit{hurts} NLI performance ($p < 0.05$).

Having established the task-dependent nature of CoT, we examine whether retrieval method choice similarly affects performance.

\subsection{BM25 vs.\ Dense Retrieval}
\label{sec:results-rag}

Our controlled ablation comparing BM25 (sparse) and dense retrieval (all-MiniLM-L6-v2) yields a striking null result: mean accuracy differs by just 0.3~pp (37.1\% vs.\ 36.8\%), with BM25 slightly ahead on \textsc{ContractNLI} ($-$0.8~pp) and \textsc{CaseHOLD} ($-$0.6~pp) while dense retrieval marginally leads on \textsc{ECtHR} (+0.5~pp).
Out of 27 paired comparisons, only 1 reaches significance at $p < 0.01$, consistent with the expected false positive rate.
This suggests BM25's lexical matching is as effective as learned dense representations for legal passage retrieval.

Detailed per-dataset model$\times$strategy breakdowns are provided in Appendix~\ref{app:detailed-results}.

\paragraph{Parse errors.}
Parse error rates (outputs unmappable to valid labels) vary substantially: Nemotron-9B produces 16.4\% unparseable outputs on average (reaching 29.8\% on \textsc{ContractNLI}), explaining much of its poor performance, while GPT-4o-mini and Qwen2.5-7B achieve near-zero rates.
Full breakdowns are in Appendix~\ref{app:parse-errors}.

\subsection{Comparison to Published Baselines}
\label{sec:baselines}

Our best prompted results remain below published fine-tuned baselines: Legal-BERT achieves 75.3\% on \textsc{CaseHOLD} with fine-tuning \citep{chalkidis-etal-2020-legal-bert} vs.\ our 71.2\% (Qwen3-A3B few-shot); SaulLM-7B reaches 83.4\% on \textsc{ContractNLI} \citep{colombo-etal-2024-saullm7b} vs.\ our 79.9\% (Haiku CoT); and fine-tuned \textsc{LexGLUE} models achieve macro-F1 above 60\% on \textsc{ECtHR} \citep{chalkidis-etal-2022-lexglue} vs.\ our 11.9\% subset accuracy.
For the two classification tasks, these gaps of 3.5--4.1~pp quantify the cost of prompting-only evaluation relative to domain-adapted fine-tuning.


\section{Conclusion}
\label{sec:conclusion}

We presented a systematic evaluation of nine language models across three legal reasoning tasks using five prompting strategies, comprising 405 experiments at a total cost of \$62.
Our central finding is that open-weight small models can match commercial APIs on legal reasoning: Qwen3-A3B (3B active parameters) matches GPT-4o-mini in mean accuracy (46.5\% vs.\ 47.2\%) and surpasses it on \textsc{CaseHOLD} with few-shot prompting (71.2\% vs.\ 67.9\%), demonstrating that architectural efficiency can substitute for raw parameter count.
Moreover, prompting strategy effectiveness is strongly task-dependent: chain-of-thought helps contract entailment ($+$8.5~pp) but hurts legal holding identification ($-$16.0~pp) and multi-label classification ($-$3.8~pp). Few-shot prompting emerges as the most consistently effective strategy.
We also find that architecture and training quality matter more than scale: a 9B dense model (Nemotron) performs worst overall while a 3B-active MoE model outperforms all dense models up to 8B, and BM25 and dense retrieval yield near-identical results (0.3~pp mean difference), suggesting the bottleneck lies in the LLM's utilization of retrieved context rather than retrieval quality.

\paragraph{Practical guidance.}
For practitioners, we recommend: (1)~Qwen3-A3B as the best cost-performance option, matching commercial APIs at open-source inference costs; (2)~few-shot prompting as the default strategy, with CoT reserved for entailment tasks; (3)~BM25 for RAG, as it matches dense retrieval without requiring an embedding model or vector database; and (4)~prioritizing well-trained 3--8B models over larger but poorly calibrated alternatives.

\paragraph{Future Work.}
Several directions merit investigation.
Comparing prompted general-purpose models against fine-tuned and domain-adapted alternatives (e.g., SaulLM; \citealt{colombo-etal-2024-saullm7b}) would clarify when fine-tuning is worth the additional investment.
Expanding to additional legal tasks, including statutory reasoning \citep{blair-stanek-etal-2023-statutory}, legal summarization, and judgment prediction, would test generalizability, as would multilingual evaluation using \textsc{LEXTREME} \citep{niklaus-etal-2023-lextreme} to assess whether our findings transfer across legal systems and languages.
Evaluating models with longer context windows on full-length legal documents (rather than our 8K-truncated inputs) may also reveal different model rankings, particularly for \textsc{ECtHR} cases that average 4.5K tokens.
Taken together, our results show that careful model and strategy selection enables open-weight small models to approach commercial API performance on legal reasoning tasks at a fraction of the cost.


\section*{Limitations}
\label{sec:limitations}

\paragraph{Prompting-only scope.}
We evaluate only prompting-based approaches without fine-tuning; parameter-efficient methods such as LoRA \citep{hu-etal-2022-lora} substantially outperform prompted models on legal benchmarks \citep{colombo-etal-2024-saullm7b}, so our results represent a lower bound.
We also use a single prompt template per strategy without per-model optimization \citep{kojima-etal-2022-zero-shot-cot}.

\paragraph{Limited scope.}
Our three English-language benchmarks cover contract law (US), case law (US), and human rights (European) but cannot represent all legal reasoning types.
We do not evaluate generative tasks (summarization, drafting), non-English jurisdictions, or civil law systems.
The \textsc{ECtHR} subset of 225 samples also limits statistical power on that benchmark.

\paragraph{Data contamination.}
We cannot rule out overlap between evaluation data and model training corpora, particularly for \textsc{CaseHOLD} (public judicial opinions) and \textsc{ContractNLI} (SEC filings).
Contamination detection is infeasible for closed-weight API models where training data composition is undisclosed.

\paragraph{Reproducibility constraints.}
All models are evaluated via cloud API endpoints, introducing dependence on provider-specific serving configurations (quantization, batching) that may differ from local deployments.
We use greedy decoding (temperature~$= 0$) and three data-sampling seeds, but API providers may update model weights without notice.
We release our evaluation harness to facilitate replication.


\section*{Ethics Statement}
\label{sec:ethics}

\paragraph{Data.}
We use three publicly available research datasets: \textsc{ContractNLI} \citep{koreeda-manning-2021-contractnli}, \textsc{CaseHOLD} \citep{zheng-etal-2021-casehold}, and \textsc{ECtHR} \citep{chalkidis-etal-2019-ecthr}, derived from published legal documents (SEC filings, judicial opinions, and court rulings, respectively).
These datasets do not contain private personal information, and their use for research purposes is consistent with their intended distribution.

\paragraph{Cost transparency and accessibility.}
The total computational cost of our 405 experiments was US\$61.88, processed entirely through commercial cloud APIs without dedicated GPU infrastructure.
We report detailed per-model and per-provider cost breakdowns to enable other researchers to estimate budgets for similar studies.
However, we acknowledge that API-based evaluation creates a dependency on commercial providers, which may present access barriers for researchers in resource-constrained settings or regions with limited API availability.
The cost disparity across providers (e.g., \$0.81 total for Gemma-3n-E4B vs.\ \$13.43 for Claude~3.5~Haiku) also means that model selection in practice may be influenced by budget constraints rather than purely by task performance.
Because we conducted all experiments via cloud APIs with no model training, the direct environmental footprint is limited to inference-time energy consumption.

\paragraph{Intended use.}
We intend this study to inform the NLP and legal AI communities about the capabilities of small language models on legal reasoning tasks.
By demonstrating that sub-10B models can approach commercial API-level performance on structured legal tasks, we hope to support efforts toward more accessible and cost-efficient legal NLP tools.
Our results should \emph{not} be interpreted as endorsements of any model for deployment in legal practice.
Legal applications of language models carry significant risks, including potential for incorrect legal analysis, hallucinated citations, and encoded biases, and require careful human oversight by qualified legal professionals.

\paragraph{Bias considerations.}
Language models may encode biases present in their training data, including biases related to jurisdiction, legal tradition, socioeconomic factors, and protected characteristics.
The \textsc{ECtHR} dataset, which concerns human rights violations, is particularly sensitive in this regard.
We evaluate models on classification accuracy but do not audit for systematic biases in model outputs across demographic or jurisdictional dimensions, which we leave as important future work.

\section*{LLM Disclosure}
Language model assistance was used during the drafting and editing process; all content, analysis, and experimental results have been verified for accuracy by the authors.
The experimental infrastructure, data collection, and statistical analysis were designed and executed by the authors.

\section*{Reproducibility Statement}
All experiments use cloud inference APIs with greedy decoding (temperature~$= 0$) and specific model identifiers listed in Table~\ref{tab:models}.
Each configuration is evaluated with three random seeds (42, 123, 456) controlling stratified data sampling, with 95\% bootstrap confidence intervals (10{,}000 resamples).
Prompt templates are in Appendix~\ref{app:prompts}.
Our evaluation harness and analysis scripts will be released upon publication.
The total cost of US\$62 (Appendix~\ref{sec:appendix-cost}) makes full replication accessible.

\bibliography{references}
\bibliographystyle{colm2026_conference}

\appendix

\section{Experimental Infrastructure and Cost}
\label{sec:appendix-cost}

Table~\ref{tab:cost-summary} provides detailed per-model inference cost and latency breakdowns for all 405 experiments across the three API providers.

\begin{table}[h]
\centering
\small
\begin{tabular}{lrrrr}
\toprule
\textbf{Model} & \textbf{Provider} & \textbf{Cost} & \textbf{Med.} & \textbf{P95} \\
 & & \textbf{(USD)} & \textbf{Lat.} & \textbf{Lat.} \\
\midrule
Haiku~3.5        & Anthropic & \$13.43 & 4.49s & 12.85s \\
Qwen2.5-7B       & Together  & \$11.05 & 1.98s & 5.52s \\
Qwen3-A3B        & Together  & \$9.56  & 2.10s & 4.32s \\
Mistral-7B       & Together  & \$9.35  & 2.26s & 3.83s \\
Llama-3.1-8B     & Together  & \$6.51  & 1.08s & 1.68s \\
GPT-4o-mini      & OpenAI    & \$5.84  & 2.78s & 6.59s \\
Nemotron-9B      & Together  & \$3.20  & 2.60s & 3.28s \\
Llama-3.2-3B     & Together  & \$2.13  & 3.10s & 8.61s \\
Gemma-3n-E4B     & Together  & \$0.81  & 5.46s & 9.99s \\
\midrule
\textbf{Total}   &           & \textbf{\$61.88} & & \\
\bottomrule
\end{tabular}
\caption{Per-model inference cost and latency across all 405 experiments (324 main + 81 ablation). Median and P95 latencies are averaged across all configurations per model. Models sorted by total cost.}
\label{tab:cost-summary}
\end{table}

All models are accessed via cloud API endpoints: Together AI\footnote{\url{https://www.together.ai}} serves the seven open-weight models, the OpenAI API serves GPT-4o-mini, and the Anthropic API serves Claude~3.5~Haiku.
The total cost of US\$61.88 covers approximately 165{,}375 individual API calls processing 314.9M input tokens and generating 26.7M output tokens.
Together AI accounts for 56\% of the total cost (\$34.62 for 315 experiments), followed by Anthropic (\$13.43 for 45 experiments) and OpenAI (\$5.84 for 45 experiments).

We use separate thread pools per API provider to maximize throughput while respecting rate limits: 29 concurrent threads for Together AI, 15 for OpenAI, and 3 for Anthropic.
The execution framework is resume-safe, skipping completed experiments based on output file existence, which proved essential when Anthropic credit depletion required re-running 23 Haiku experiments with a fresh API key.

Dense retrieval embeddings are computed using the \texttt{all-MiniLM-L6-v2} sentence-transformer model and cached to disk, adding negligible cost to the retrieval ablation experiments.

\section{Prompting Strategy Templates}
\label{app:prompts}

\begin{figure}[h]
\centering
\includegraphics[width=\columnwidth]{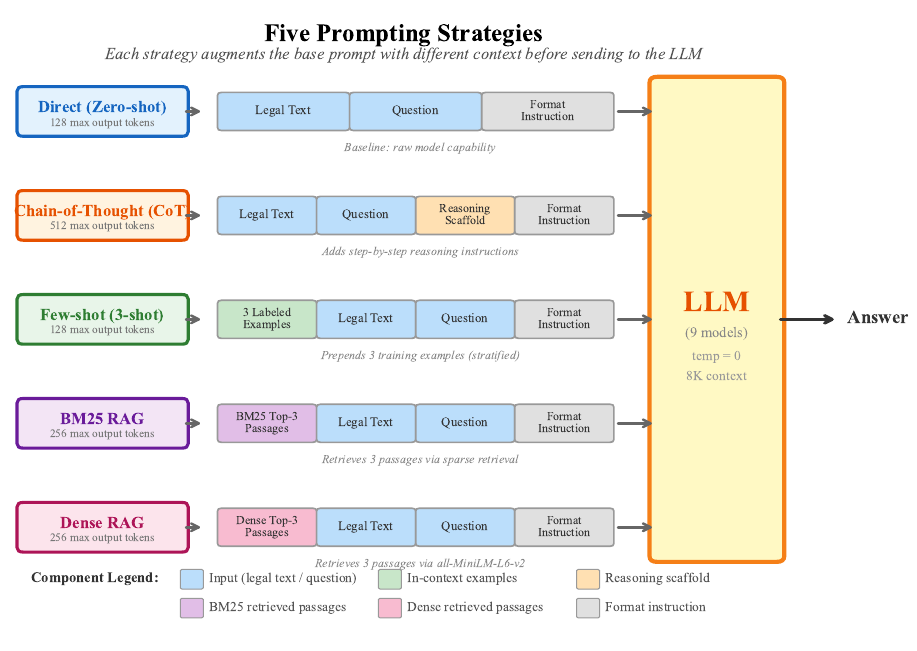}
\caption{Overview of the five prompting strategies. Each row shows the prompt components assembled for a given strategy. All strategies share the core components (legal text, question, format instruction) but differ in the additional context provided. Output token budgets vary by strategy: 128 tokens (direct, few-shot), 256 tokens (RAG), and 512 tokens (CoT).}
\label{fig:prompting-strategies}
\end{figure}

Figure~\ref{fig:prompting-strategies} provides a visual overview of the five prompting strategies evaluated in this study. Each strategy augments the base prompt with different context components before sending it to the model. The direct strategy serves as the baseline, while the remaining strategies add reasoning scaffolds (CoT), in-context examples (few-shot), or retrieved passages (BM25 and dense RAG).

\section{Detailed Per-Dataset Results}
\label{app:detailed-results}

Tables~\ref{tab:detail-cnli}--\ref{tab:detail-ecthr} provide the full model$\times$strategy accuracy for each dataset. Model names are abbreviated; see Table~\ref{tab:models} for full names.

\begin{table}[h]
\centering
\small
\setlength{\tabcolsep}{3pt}
\begin{tabular}{lccccc}
\toprule
\textbf{Model} & \textbf{Dir} & \textbf{FS} & \textbf{CoT} & \textbf{R\textsubscript{B}} & \textbf{R\textsubscript{D}} \\
\midrule
GPT-4o-mini    & \underline{.761} & \textbf{.791} & .771 & .754 & .755 \\
Qwen3-A3B      & .759 & \underline{.767} & \underline{.776} & \textbf{.787} & \textbf{.785} \\
Haiku          & .657 & .618 & \textbf{.799} & .638 & .645 \\
Mistral-7B     & .675 & .678 & .659 & \underline{.723} & \underline{.696} \\
Qwen2.5-7B     & .657 & .697 & .680 & .688 & .691 \\
Gemma-3n-E4B   & .605 & .719 & .687 & .678 & .655 \\
Llama-3.1-8B   & .641 & .608 & .544 & .558 & .539 \\
Llama-3.2-3B   & .273 & .421 & .432 & .474 & .447 \\
Nemotron       & .101 & .175 & .544 & .343 & .360 \\
\bottomrule
\end{tabular}
\caption{\textsc{ContractNLI} accuracy (mean over 3 seeds). Dir = Direct, FS = Few-Shot, R\textsubscript{B} = RAG\textsubscript{BM25}, R\textsubscript{D} = RAG\textsubscript{Dense}. Best in \textbf{bold}, second-best \underline{underlined}.}
\label{tab:detail-cnli}
\end{table}

\begin{table}[h]
\centering
\small
\setlength{\tabcolsep}{3pt}
\begin{tabular}{lccccc}
\toprule
\textbf{Model} & \textbf{Dir} & \textbf{FS} & \textbf{CoT} & \textbf{R\textsubscript{B}} & \textbf{R\textsubscript{D}} \\
\midrule
Qwen3-A3B      & \textbf{.692} & \textbf{.712} & .373 & .498 & .513 \\
GPT-4o-mini    & \underline{.681} & \underline{.679} & .499 & \textbf{.634} & \textbf{.615} \\
Qwen2.5-7B     & .649 & .649 & .277 & \underline{.583} & .563 \\
Mistral-7B     & .606 & .619 & .342 & .540 & .511 \\
Llama-3.1-8B   & .618 & .595 & \underline{.535} & .561 & \underline{.579} \\
Llama-3.2-3B   & .577 & .482 & .371 & .407 & .415 \\
Haiku          & .343 & .281 & \textbf{.537} & .285 & .275 \\
Gemma-3n-E4B   & .464 & .575 & .261 & .381 & .374 \\
Nemotron       & .203 & .199 & .201 & .211 & .205 \\
\bottomrule
\end{tabular}
\caption{\textsc{CaseHOLD} accuracy (mean over 3 seeds). Best in \textbf{bold}, second-best \underline{underlined}. Note the consistent CoT degradation across models, with Haiku as the sole exception.}
\label{tab:detail-casehold}
\end{table}

\begin{table}[h]
\centering
\small
\setlength{\tabcolsep}{3pt}
\begin{tabular}{lccccc}
\toprule
\textbf{Model} & \textbf{Dir} & \textbf{FS} & \textbf{CoT} & \textbf{R\textsubscript{B}} & \textbf{R\textsubscript{D}} \\
\midrule
Haiku          & \textbf{.073} & \textbf{.119} & .000 & .016 & .004 \\
Qwen3-A3B      & .059 & \underline{.085} & \textbf{.031} & \textbf{.074} & \underline{.064} \\
Llama-3.1-8B   & .058 & .073 & .018 & .024 & .024 \\
Qwen2.5-7B     & .050 & .059 & \underline{.019} & \underline{.062} & \textbf{.065} \\
Mistral-7B     & \underline{.061} & .022 & .006 & .019 & .025 \\
Gemma-3n-E4B   & .052 & .025 & .000 & .028 & .042 \\
GPT-4o-mini    & .050 & .028 & .004 & .024 & .039 \\
Nemotron       & .028 & .042 & .007 & .013 & .027 \\
Llama-3.2-3B   & .010 & .006 & .013 & .019 & .037 \\
\bottomrule
\end{tabular}
\caption{\textsc{ECtHR} subset accuracy (mean over 3 seeds, $n$=225 per experiment). Best in \textbf{bold}, second-best \underline{underlined}. Haiku few-shot achieves 11.9\%, the only configuration exceeding 10\%.}
\label{tab:detail-ecthr}
\end{table}

\section{Parse Error Analysis}
\label{app:parse-errors}

Table~\ref{tab:parse-errors} reports the proportion of model outputs that could not be mapped to valid answer labels.

\begin{table}[h]
\centering
\small
\begin{tabular}{lccc|c}
\toprule
\textbf{Model} & \textbf{\textsc{CNLI}} & \textbf{\textsc{CH}} & \textbf{\textsc{ECtHR}} & \textbf{Mean} \\
\midrule
GPT-4o-mini    & .000 & .000 & .001 & .000 \\
Qwen2.5-7B     & .000 & .000 & .001 & .000 \\
Mistral-7B     & .003 & .000 & .002 & .002 \\
Llama-3.1-8B   & .003 & .000 & .007 & .003 \\
Llama-3.2-3B   & .001 & .002 & .010 & .004 \\
Qwen3-A3B      & .001 & .004 & .035 & .013 \\
Gemma-3n-E4B   & .001 & .005 & .079 & .028 \\
Haiku          & .132 & \textbf{.006} & .038 & .059 \\
Nemotron       & \textbf{.298}$^\dagger$ & .004 & \textbf{.189}$^\dagger$ & \textbf{.164}$^\dagger$ \\
\bottomrule
\end{tabular}
\caption{Mean parse error rate by model and dataset (proportion of unparseable outputs). Worst (highest) per column in \textbf{bold}. $^\dagger$Rates exceeding 10\%.}
\label{tab:parse-errors}
\end{table}

Nemotron-9B exhibits the highest parse error rates (16.4\% overall, reaching 29.8\% on \textsc{ContractNLI}), explaining much of its poor performance.
Haiku shows elevated \textsc{ContractNLI} parse errors (13.2\%).
GPT-4o-mini and Qwen2.5-7B achieve near-zero parse error rates across all conditions, indicating strong instruction-following capabilities.

\end{document}